\documentclass[runningheads]{llncs}
\usepackage[T1]{fontenc}
\usepackage{microtype}
\usepackage{graphicx}
\usepackage{subfigure}
\usepackage{booktabs} 
\usepackage{xcolor}

\usepackage{amsmath}
\usepackage{amssymb}
\usepackage{mathtools}

\begin{document}
%
\title{Fusing Modalities by Multiplexed Graph Neural Networks for Outcome Prediction in Tuberculosis}
%
%
\author{Niharika S. D'Souza\inst{1}  \and Hongzhi Wang \inst{1} \and Andrea Giovannini \inst{2} \and Antonio Foncubierta-Rodriguez \inst{2} \and Kristen L. Beck \inst{1} \and Orest Boyko \inst{3} \and Tanveer Syeda-Mahmood \inst{1}}
%

\authorrunning{N.S. D'Souza et al.}
%
\institute{IBM Research Almaden, San Jose, CA, USA \and IBM Research, Zurich, Switzerland \and Department of Radiology,
VA Southern Nevada Healthcare System, NV, USA}

\maketitle              
\begin{abstract}
In a complex disease such as tuberculosis, the evidence for the disease and its evolution may be present in multiple modalities such as clinical, genomic, or imaging data. Effective patient-tailored outcome prediction and therapeutic guidance will require fusing evidence from these modalities. Such multimodal fusion is difficult since the evidence for the disease may not be uniform across all modalities, not all modality features may be relevant, or not all modalities may be present for all patients. All these nuances make simple methods of early, late, or intermediate fusion of features inadequate for outcome prediction. In this paper, we present a novel fusion framework using multiplexed graphs and derive a new graph neural network for learning from such graphs. Specifically, the framework allows modalities to be represented through their targeted encodings, and models their relationship explicitly via multiplexed graphs derived from salient features in a combined latent space. We present results that show that our proposed method outperforms state-of-the-art methods of fusing modalities for multi-outcome prediction on a large Tuberculosis (TB) dataset. 
\end{abstract}

\section{Introduction}
Tuberculosis (TB) is one of the most common infectious diseases worldwide \cite{world2010treatment}. Although the mortality rate caused by TB has declined in recent years, single- and multi-drug resistance has become a major threat to quick and effective TB treatment. Studies have revealed that predicting the outcome of a treatment is a function of many patient-specific factors for which collection of multimodal data covering clinical, genomic, and imaging information about the patient has become essential \cite{munoz2010factors}.  However, it is not clear what information is best captured in each modality and how best to combine them. For example, genomic data can reveal the genetic underpinnings of drug resistance and identify genes/mutations conferring drug-resistance \cite{manson2017genomic}. Although imaging data from X-ray or CT can show statistical difference for drug resistance, they alone may be insufficient to differentiate multi-drug resistant TB from drug-sensitive TB \cite{wang2018radiological}. Thus it is important to develop methods that allow simultaneously extraction of relevant information from multiple modalities as well as ways to combine them in an optimal fashion to lead to better outcome prediction.

Recent efforts to study the fusion problem for outcome prediction in TB have focused on a single outcome such as treatment failure or used only a limited number of modalities such as clinical and demographic data \cite{sauer2018feature,asad2020machine}. In this paper, we take a comprehensive approach by treating the outcome prediction problem as a multiclass classification for multiple possible outcomes through multimodal fusion. We leverage more extensive modalities beyond clinical, imaging, or genomic data, including novel features extracted via advanced analysis of protein domains from genomic sequence as well as deep learning-derived features from CT images as shown in Fig.~\ref{TB Results}.(a). Specifically, we develop a novel fusion framework using multiplexed graphs to capture the information from modalities and derive a new graph neural network for learning from such graphs. The framework represents modalities through their targeted encodings, and models their relationship via multiplexed graphs derived from projections in a latent space.

Existing approaches often infer matrix or tensor encodings from individual modalities ~\cite{lahat2015multimodal} combined with early, late, or intermediate fusion~\cite{subramanian2020multimodal,baltruvsaitis2018multimodal,Wang2021} of the individual representations. Example applications include- CCA for speaker identification~\cite{sargin2006multimodal}, autoencoders for video analytics~\cite{vu2017multimodal}, transformers for VQA~\cite{kant2020spatially}, etc. In contrast, our approach allows for the modalities to retain their individuality while still participating in exploring explicit relationships between the modality features through the multiplexed framework. Specifically, we design our framework to explicitly model relationships within and across modality features via a self-supervised multi-graph construction and design a novel graph neural network for reasoning from these feature dependencies via structured message passing walks. We present results which show that by relaxing the fusing constraints through the multiplex formulation, our method outperforms state-of-the-art methods of multimodal fusion in the context of multi-outcome prediction for TB treatments. 

\section{A Graph Based Multimodal Fusion Framework}
\begin{figure}[t!]
\begin{center}
\centerline{{\includegraphics[scale= 0.25]{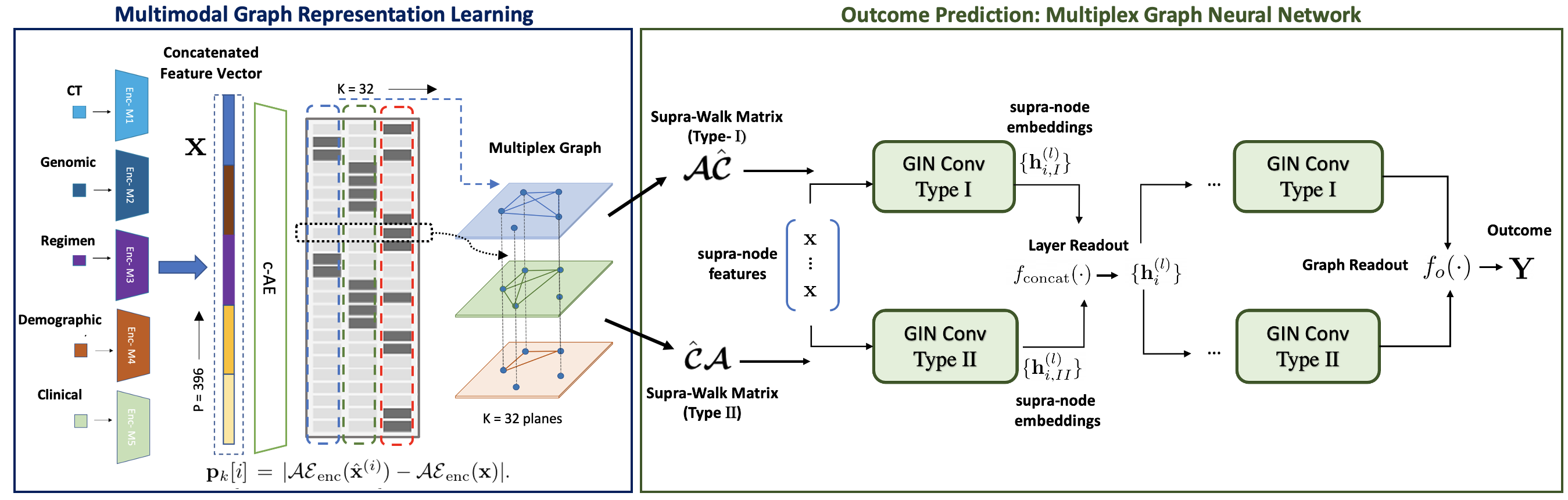}}}
\caption{\small{Graph Based Multimodal Fusion for Outcome Prediction. \textbf{Blue Box:} Incoming modality features are concatenated into a feature vector (of size P=396) and projected into a common latent space (of size K=32). Salient activations in the latent space are used to form the planes of the multiplexed graph. \textbf{Green Box:} The multiplexed GNN uses message passing walks to combine latent concepts for inference.}}
\label{Multiplex_Formulation}
\end{center}
\end{figure}

As alluded to earlier, exploring various facets of cross-modal interactions is at the heart of the multimodal fusion problem. To this end, we propose to utilize the representation learning theory of multiplexed graphs to develop a generalized framework for multimodal fusion. A multiplexed graph~\cite{Cozzo2018} is a type of multigraph in which the nodes are grouped into multiple planes, each representing an individual edge-type. The information captured within a plane is multiplexed to other planes through diagonal connections as shown in Fig.~\ref{Multiplex_Formulation}. Mathematically, we define a multiplexed graph as: $\mathcal{G}_{\text{Mplex}} = (\mathcal{V}_{\text{Mplex}},\mathcal{E}_{\text{Mplex}})$, where $\vert{\mathcal{V}_{\text{Mplex}}}\vert= \vert{\mathcal{V}}\vert \times K$ and $\mathcal{E}_{\text{Mplex}} = \{(i,j) \in \mathcal{V}_{\text{Mplex}} \times \mathcal{V}_{\text{Mplex}}\}$. There are $K$ distinct types of edges which can link two given nodes. Analogous to ordinary graphs, we have $k$ adjacency matrices $\mathbf{A}_{(k)} \in \mathcal{R}^{P \times P}$, where $P=\vert{\mathcal{V}}\vert$, each summarizing the connectivity information given by the edge-type $k$. The elements of these matrices are binary $\mathbf{A}_{(k)}[m,n] = 1 $ if there is an edge of type $k$ between nodes $m,n \in \mathcal{V}$. 

\paragraph{\textbf{Multimodal Graph Representation Learning:}}  While the multiplexed graph has been used for various modeling purposes in literature~\cite{kivela,manlio,Ferriani2013,Maggioni2013}, we propose to use it for multimodal fusion of imaging, genomic and clinical data for outcome prediction in TB. We adopt the construction shown in the Blue Box in Fig.~\ref{Multiplex_Formulation} to produce the multiplexed graph from the individual modality features. First, domain specific autoencoders (d-AE) are used to convert each modality into a compact feature space that can provide good reconstruction using Mean Squared Error (MSE). To capture feature dependencies across modalities, the concatenated features are brought to a common low dimensional subspace through a common autoencoder (c-AE) trained to reconstruct the concatenated features. Each latent dimension of the autoencoder captures an abstract aspect of the multimodal fusion problem, e.g. features projected to be salient in the same latent dimension are likely to form meaningful joint patterns for a specific task, and form a ``conceptual'' plane of the multiplexed graph. The $\vert{\mathcal{V_{\text{Mplex}}}}\vert$ ``supra-nodes'' of $\mathcal{G}_{\text{Mplex}}$ are produced by creating copies of features (i.e. nodes) across the planes. The edges between nodes in each plane represent features whose projections in the respective latent dimensions were salient (see section \ref{exp:graph_construction} for details). Further, each plane is endowed with its own topology and is a proxy for the correlation between features across the corresponding latent dimension. This procedure helps model the interactions between the various modality features in a principled fashion. We thus connect supra-nodes within a plane to each other via the intra-planar adjacency matrix $\mathbf{A}_{(k)}$, allowing us to traverse the multi-graph according to the edge-type $k$. We also connect each supra-node with its own copy in other planes via diagonal connections, allowing for inter-planar traversal.

\paragraph{\textbf{Outcome Prediction via the Multiplexed GNN:}}
We develop a novel graph neural network for outcome prediction from the multiplexed graph (Green Box in Fig.~\ref{Multiplex_Formulation}). Graph Neural Networks (GNN) are a class of representation learning algorithms that distill connectivity information to guide a downstream inference task \cite{scarselli2008graph}. A typical GNN schema comprises of two components: (1) a message passing scheme for propagating information across the graph and (2) task-specific supervision to guide the representation learning. For ordinary graphs, the adjacency matrix $\mathbf{A}$ and its matrix powers allows us to keep track of neighborhoods (at arbitrary $l$ hop distance) within the graph during message passing. Conceptually, cascading $l$ GNN layers is analogous to pooling information at each node $i$ from its $l$-hop neighbors that can be reached by a walk starting at $i$. The Multiplex GNN is designed to mirrors this behavior.

The \textit{intra-planar adjacency matrix} $\boldsymbol{\mathcal{A}}\in \mathcal{R}^{PK \times PK} $, and the \textit{inter-planar transition control matrix} $\hat{\boldsymbol{\mathcal{C}}} \in \mathcal{R}^{PK \times PK}$ \cite{Cozzo2018} define walks on the multiplex $\mathcal{G}_{\text{Mplex}}$. 
\begin{equation}
    \boldsymbol{\mathcal{A}} = \bigoplus_{k}\mathbf{A}_{(k)} \ \ \ \ \  ; \ \ \ \ \ 
    \hat{\boldsymbol{\mathcal{C}}} = [\mathbf{1}_{K}\mathbf{1}_{K}^{T}] \otimes \boldsymbol{\mathcal{I}}_{P}   \label{ILtrans}
\end{equation}
where $\bigoplus$ is the direct sum operation, $\otimes$ denotes the Kronecker product, $\mathbf{1}_{K}$ is the $K$ vector of all ones, and $\boldsymbol{\mathcal{I}}_{P}$ denotes the identity matrix of size $P \times P$. Thus $\boldsymbol{\mathcal{A}}$ is block-diagonal by construction and captures within plane transitions across supra-nodes. Conversely, $\hat{\boldsymbol{\mathcal{C}}}$ has identity matrices along on off-diagonal blocks. This implicitly restricts across plane transitions to be between supra-nodes which arise from the same multi-graph node (i.e. $i$ and $P(k-1)+i$ for $k \in \{1,\dots, K\}$). Since supra-nodes across planes can already be reached by combining within and across-planar transitions, this provides comparable representational properties at a reduced complexity ($\mathcal{O}(PK)$) inter-planar edges instead of $\mathcal{O}(P^2K)$).

A walk on $ \mathcal{G}_{\text{Mplex}}$ combines within and across planar transitions to reach a supra-node $j\in \mathcal{V}_{\text{Mplex}}$ from a given supra-node $i \in \mathcal{V}_{\text{Mplex}}$. $\boldsymbol{\mathcal{A}}$ and $\hat{\boldsymbol{\mathcal{C}}}$ allow us to define multi-hop transitions on the multiplex in a convenient factorized form. A multiplex walk proceeds according to two types of transitions \cite{Cozzo2018}: (1) A single intra-planar step or (2) A step that includes both an inter-planar step moving from one plane to another (this can be before or after the occurrence of an intra-planar step). To recreate these transitions exhaustively, we have two supra-walk matrices. $\boldsymbol{\mathcal{A}}\hat{\boldsymbol{\mathcal{C}}}$ encodes transitions where \textit{after} an intra-planar step, the walk \textit{can} continue in the same plane or transition to a different plane (Type I). Similarly, using $\hat{\boldsymbol{\mathcal{C}}}\boldsymbol{\mathcal{A}}$, the walk \textit{can} continue in the same plane or transition to a different plane \textit{before} an intra-planar step (Type II).

\paragraph{\textbf{Message Passing Walks:}} Let $\mathbf{h}^{l}_{i} \in \mathcal{R}^{D^{l}\times 1}$ denote the (supra)-node representation for (supra)-node $i$. In matrix form, we can write $\mathbf{H}^{(l)} \in \mathcal{R}^{\vert{\mathcal{V}_{\text{Mplex}}}\vert \times D^{l}}$, with $\mathbf{H}^{(l)}[i,:] = \mathbf{h}^{(l)}_{i}$. We then compute this via the following operations:
\begin{eqnarray}
\mathbf{h}_{i,I}^{(l+1)} = \boldsymbol{\phi}_{I}\Big(\{\mathbf{h}^{(l)}_{j}, j: [\boldsymbol{\mathcal{A}}\hat{\boldsymbol{\mathcal{C}}}][i,j] = 1 \}\Big) \ \  ;  \ \ \ 
\mathbf{h}_{i,II}^{(l+1)} = \boldsymbol{\phi}_{II}\Big(\{\mathbf{h}^{(l)}_{j}, j: [\hat{\boldsymbol{\mathcal{C}}}\boldsymbol{\mathcal{A}}][i,j] = 1 \}\Big)  \nonumber \\ 
\mathbf{h}^{(l+1)}_{i} = f_{\text{concat}}(\mathbf{h}^{(l+1)}_{i,I},\mathbf{h}^{(l+1)}_{i,II}) \ \ \ \ \ ; \ \ \ \ \  f_{\text{o}}(\{\mathbf{h}^{(L)}_{i}\}) = \mathbf{Y} \ \ \ \ \ \ \ \ \ \label{Mplex_MP}
\end{eqnarray}
Here, $f_{\text{concat}}(\cdot)$ concatenates the Type I and Type II representations. At the input layer, we have $\mathbf{H}^{(0)} = \mathbf{X} \otimes \mathbf{1}_{K}$, where $\mathbf{X} \in \mathcal{R}^{\vert{V}\vert \times 1}$ are the node inputs (concatenated modality features). $\{\boldsymbol{\mathcal{\phi}}_{I}(\cdot), \boldsymbol{\mathcal{\phi}}_{II}(\cdot)\}$ performs message passing according to the neighborhood relationships given by the supra-walk matrices. Finally, $f_{o}(\cdot)$ is the graph readout that predicts the outcome $\mathbf{Y}$. The learnable parameters of the Multiplex GNN can be estimated via standard backpropagation

\paragraph{\textbf{Implementation Details:}}
We utilize the Graph Isomorphism Network (GIN) \cite{xu2018powerful} with LeakyReLU (neg. slope = 0.01) readout for message passing (i.e. $\{\boldsymbol{\mathcal{\phi}}_{I}(\cdot), \boldsymbol{\mathcal{\phi}}_{II}(\cdot)\}$). Since the input $\mathbf{x}$ is one dimensional, we have two such layers in cascade with hidden layer width one. $f_{o}(\cdot)$ is a Multi-Layered Perceptron (MLP) with two hidden layers (size: 100 and 20) and LeakyReLU activation. In each experimental comparison, we chose the model architecture and hyperparameters for our framework (learning rate=0.001 decayed by 0.1 every 20 epochs, weight decay=0.001, number of epochs =40) and baselines using grid-search and validation set. All frameworks are trained on the Cross Entropy loss between the predicted logits (after a softmax) and the ground truth labels. We utilize the ADAMw optimizer \cite{loshchilov2017decoupled}. Models were implemented using the Deep Graph Library (v=0.6.2) in PyTorch (v=0.10.1). We trained all models on a 64GB CPU RAM, 2.3 GHz 8-Core Intel i9 machine, with 3.5-4 hrs training time per run (Note: Performing inference via GPUs will likely speed up computation).

\section{Experimental Evaluation}
\begin{table}[b!]
\footnotesize{
\begin{center}
\caption{Dataset Description of the TB dataset. }
\begin{tabular}{|c|c|c|c|c|c|c|}
\hline
\textbf{Modality}& \textbf{CT} & \textbf{Genomic}&\textbf{Demographic}& \textbf{Clinical}&\textbf{Regimen}& \textbf{Continuous}\\ 
\hline
Native Dimen. & 2048  & 4081 &  29 & 1726 & 233 & 8\\
Rank & 250  & 300 &  24 & 183 & 112 & 8\\
Reduced Dim. & 128  & 64 &  8 & 128 & 64 & 4\\
\hline
\end{tabular}
\label{Dataset Description}
\end{center}}
\end{table}
\subsection{Data and Experimental Setup}
\label{Data}
We conducted experiments using the Tuberculosis Data Exploration Portal \cite{Gabrielian2019}. 3051 patients with five classes of treatment outcomes (Still on treatment, Died, Cured, Completed, or Failure) were used. Five modalities are available (see Fig. \ref{TB Results}.(a)). Demographic, clinical, regimen and genomic data are available for each patient, while chest CTs are available for 1015 patients. For clinical and regimen data, information that might be directly related to treatment outcomes, such as type of resistance, were removed. For each CT, lung was segmented using multi-atlas segmentation \cite{wang2013multi}. The pre-trained DenseNet \cite{huang2017densely} was then applied to extract a feature vector of 1024-dimension for each axial slice intersecting lung. To aggregate the information from all lung intersecting slices, the mean and maximum of each of the 1024 features were used providing a total of 2048 features. For genomic data from the causative organisms \textit{Mycobacterium tuberculosis} (Mtb), 81 single nucleotide polymorphisms (SNPs) in genes known to be related to drug resistance were used. In addition, we retrieved the raw genome sequence from NCBI Sequence Read Archive for 275 patients to describe the biological sequences of the disease-causing pathogen at a finer granularity. The data was processed by the IBM Functional Genomics Platform \cite{Seabolt2019}. Briefly, each Mtb genome underwent an iterative \textit{de novo} assembly process and then processed to yield gene and protein sequences. The protein sequences were then processed using InterProScan \cite{Jones2014} to generate the functional domains. Functional domains are sub-sequences located within the protein's amino acid chain. They are responsible for the enzymatic bioactivity of a protein and can more aptly describe the protein's function. 4000 functional features were generated for each patient. 

{\noindent\textbf{Multiplexed Graph Construction:}
\label{exp:graph_construction}
We note that the regimen and genomic data are categorical features. CT features are continuous. The demographic and clinical data are a mixture of categorical and continuous features. Grouping the continuous demographic and clinical variables together yielded a total of six source modalities (see Table~\ref{Dataset Description}). We impute the missing CT and functional genomic features using the mean values from the training set. To reduce the redundancy in each domain, we use d-AEs with fully connected layers, LeakyReLU non-linearities and tied weights trained to reconstruct the raw modality features. The d-AE bottleneck (see Table~\ref{Dataset Description}) is chosen via the validation set.
The reduced individual modality features are concatenated to form the node feature vector $\mathbf{x}$. To form the multiplexed graph planes, the c-AE projects $\mathbf{x}$ to a `conceptual' latent space of dimension $K<<P$ where $P=128+64+8+128+64+4 = 396$. We use the c-AE concept space form the planes of the multiplex and explore the correlation between pairs of features. The c-AE architecture mirrors the d-AE, but projects the training examples $\{\mathbf{x}\}$ to $K=32$ concepts. We infer within plane connectivity along each concept perturbing the features and recording those features giving rise to largest incremental responses. Let $\mathcal{AE}_{\text{enc}}(\cdot): \mathcal{R}^{P} \rightarrow \mathcal{R}^{K}$ be the c-AE mapping to the concept space. Let $\hat{\mathbf{x}}^{(i)}$ denote the perturbation of the input by setting $\hat{\mathbf{x}}^{(i)}[j] = \mathbf{x}[j] \ \forall \ j \neq i$ and 0 for $j=i$. Then for concept axis $k$, the perturbations are $\mathbf{p}_{k}[i] = \vert{\mathcal{AE}_{\text{enc}}(\hat{\mathbf{x}}^{(i)})- \mathcal{AE}_{\text{enc}}({\mathbf{x}})}\vert$. Thresholding $\mathbf{p}_{k} \in\mathcal{R}^{P \times 1}$ selects feature nodes with the strongest responses along concept $k$. To encourage sparsity, we retain the top one percent of salient patterns. We connect all pairs of such feature nodes with edge-type $k$ via a fully connected (complete) subgraph between nodes thus selected (Fig.~\ref{Multiplex_Formulation}). Across the $K$ concepts, we expect that different sets of features are prominent.  The input features $\mathbf{x}$ are one dimensional node embeddings (or the messages at input layer $l=0$).} The latent concepts $K$, and the feature selection (sparsity) are key quantities that control generalization.

\subsection{Baselines}
We compared with four multimodal fusion approaches. We also present three ablations, allowing us to probe the inference (Multiplex GNN) and representation learning (Multimodal Graph Construction) separately.

\noindent \textbf{No Fusion:} This baseline utilizes a two layered MLP (hidden width: 400 and 20, LeakyReLU activation) on the individual modality features before the d-AE dimensionality reduction. This provides a benchmark for the outcome prediction performance of each modality separately.

\noindent \textbf{Early Fusion:} 
Individual modalities are concatenated before dimensionality reduction and fed through the same MLP architecture as described above.

\noindent \textbf{Intermediate Fusion:} In this comparison, we perform intermediate fusion after the d-AE projection by using the concatenated feature $\mathbf{x}$ as input to a two layered MLP (hidden width: 150 and 20, LeakyReLU activation). This helps us evaluate the benefit of using graph based fusion via the c-AE latent encoder.

\noindent \textbf{Late Fusion:} We utilize the late fusion framework of \cite{wang2021modeling} to combine the predictions from the modalities trained individually in the No Fusion baseline. This framework leverages the uncertainty in the 6 individual classifiers to improve the robustness of outcome prediction. We used the hyperparameters in \cite{wang2021modeling}.

\noindent \textbf{Relational GCN on a Multiplexed Graph:} This baseline utilizes the multigraph representation learning (Blue Box of Fig.~\ref{Multiplex_Formulation}), but replaces the Multiplex GNN feature extraction with the Relational GCN framework of \cite{schlichtkrull2018modeling}. Essentially, at each GNN layer, the RGCN runs $K$ separate message passing operations on the planes of the multigraph and then aggregates the messages post-hoc. Since the width, depth and graph readout is the same as with the Multiplex GNN, this helps evaluate the expressive power of the walk based message passing in Eq.~(\ref{Mplex_MP}).

\noindent \textbf{Relational GCN w/o Latent Encoder:} For this comparison, we utilize the reduced features after the d-AE, but instead create a multi-layered graph with the individual modalities in different planes. Within each plane, nodes are fully connected to each other after which a two layered RGCN \cite{schlichtkrull2018modeling} model is trained. Effectively, \textit{within modality} feature dependence may still be captured in the planes, but the concept space is not used to infer the \textit{cross-modal} interactions.
\noindent \textbf{GCN on monoplex feature graph:} This baseline also incorporates a graph based representation, but does not include the use of latent concepts to model within and cross-modal feature correlations. Essentially, we construct a fully connected graph on $\mathbf{x}$ instead of using the (multi-) conceptual c-AE space and train a two layered Graph Convolutional Network \cite{kipf2016semi} for outcome prediction.

\subsection{Results}

\paragraph{\textbf{Evaluation Metrics:}} Since we have unbalanced classes for multi-class classification, we evaluate the performance using AU-ROC (Area Under the Receiver Operating Curve). We also report weighted average AU-ROC as an overall summary. We rely on $10$ randomly generated train/validation/test splits of size 2135/305/611 to train the representation learning and GNNs in a fully blind fashion. We use the same splits and training/evaluation procedure for the baselines. For statistical rigour, we indicate significant differences between Multiplex GNN and baseline AU-ROC for each class as quantified by a DeLong \cite{delong1988comparing} test.
\begin{figure}[t!]
\begin{center}
\centerline{\includegraphics[scale=0.38]{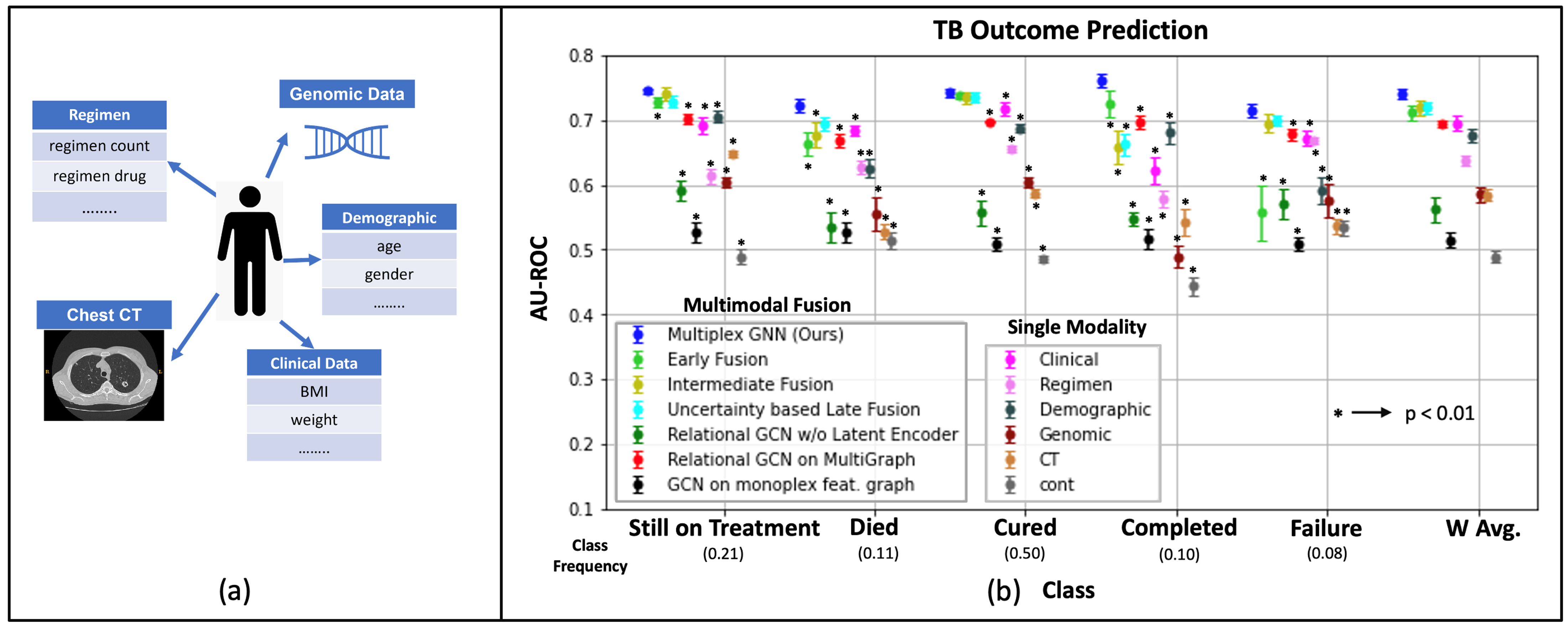}}
\caption{(a). Multimodal data for Tuberculosis treatment outcome prediction. (b). Outcome prediction performance measured by per-class and weighted average AU-ROC. We display mean performance along with standard errors. * indicates comparisons with the Multiplexed GNN per-class AU-ROC with ($p<0.01$) according to the DeLong test. Individual class frequencies are listed below the x axis.
}
\label{TB Results}
\end{center}
\end{figure}
\paragraph{\textbf{Outcome Prediction Performance:}}
Fig.~\ref{TB Results} illustrates the outcome prediction results. Our framework outperforms common multimodal fusion baselines (Early Fusion, Intermediate Fusion, and Late Fusion), as quantified by the higher mean per-class AU-ROC and weighted average AU-ROC. Our graph based multimodal fusion also provides improved performance over the single modality outcome classifiers. The Relational GCN- Multigraph baseline is an ablation that replaces the Multiplexed GNN with an existing state-of-the art GNN framework. For both techniques, we utilize the same Multi-Graph representation as learned by the c-AE autoencoder latent space. The performance gains within this comparison suggest that the Multiplexed GNN is better suited for reasoning and task-specific knowledge distillation from multigraphs. We conjecture that the added representational power is a direct consequence of our novel multiplex GNN message passing (Eq.~(\ref{Mplex_MP})) scheme. Along similar lines, the Relational GCN w/o latent encoder and the GCN on the monoplex feature graph baseline comparisons are generic graph based fusion approaches. They allow us to examine the benefit of using the salient activation patterns from the c-AE latent concept space to infer the multi-graph representation. Specifically, the former separates the modality features into plane-specific fully connected graphs within a multi-planar representation. The latter constructs a single fully connected graph on the concatenated modality features. Our framework provides large gains over these baselines. In turn, this highlights the efficacy of our Multimodal graph construction. We surmise that the salient learned conceptual patterns are more successful at uncovering cross modal interactions between features that are explanative of patient outcomes. Overall, these observations highlight key representational aspects of our framework, and demonstrate the efficacy for the TB outcome prediction task. Given the clinical relevance, a promising direction for exploration would be to extend frameworks for explainability in GNNs (for example, via subgraph exploration~\cite{yuan2021explainability}) to Multiplex GNNs to automatically highlight patterns relevant to downstream prediction.

\section{Conclusion}

We have introduced a novel Graph Based Multimodal Fusion framework to combine imaging, genomic and clinical data. Our Multimodal Graph Representation Learning projects the individual modality features into abstract concept spaces, wherein complex cross modal dependencies can be mined from the salient patterns. We developed a new multiplexedh Neural Network that can track information flow within the multi-graph via message passing walks. Our GNN formulation provides the necessary flexibility to mine rich representations from multimodal data. Overall, this provides for improved Tuberculosis outcome prediction performance against several state-of-the-art baselines. 
\bibliographystyle{splncs04}
\bibliography{paper2619_refs.bib}
\end{document}